\documentclass[letterpaper, 10 pt, journal, twoside]{ieeetran}

\IEEEoverridecommandlockouts                              

\usepackage{graphics} 
\usepackage{epsfig} 
\usepackage{times} 
\usepackage{amsmath} 
\usepackage{amssymb}  

\usepackage{cite}
\usepackage{graphicx}
\usepackage{caption}
\usepackage{subcaption}
\usepackage[ruled, linesnumbered]{algorithm2e}
\SetAlFnt{\small}

\SetCommentSty{mycommfont}
\SetKwInput{KwInput}{Input}                
\SetKwInput{KwOutput}{Output}              

\usepackage{pdfpages}
\usepackage[font=small,labelfont=bf]{caption}
\usepackage{array}
\usepackage{booktabs}
\usepackage{diagbox}
\usepackage{stfloats}
\usepackage{todonotes}

\makeatletter
\let\NAT@parse\undefined
\makeatother
\usepackage[backref=page]{hyperref}
\hypersetup{
     colorlinks = true,
     linkcolor = green,
     anchorcolor = green,
     citecolor = green,
     filecolor = green,
     urlcolor = cyan
 }
\title{
Off-Policy Evaluation with Online Adaptation for Robot Exploration in
Challenging Environments
}

\author{Yafei Hu$^{1}$, 
Junyi Geng$^{1,2}$,
Chen Wang$^{1,3}$, 
John Keller$^{1}$,
and Sebastian Scherer$^{1}$
\thanks{$^{1}$ The Robotics Institute, Carnegie Mellon University, Pittsburgh, PA 15213, USA.
        {\tt\small \{yafeih, jkeller2, basti\}@andrew.cmu.edu}
        }%
\thanks{$^{2}$ Department of Aerospace Engineering, Pennsylvania State University, University Park, PA, 16802, USA.
        {\tt\small jgeng@psu.edu}
        }%
\thanks{$^{3}$ Department of Computer Science and Engineering, State University of New York at Buffalo, USA.
        {\tt\small chenwang@dr.com}
        }%
}

\begin{document}

\maketitle



\begin{abstract}
Autonomous exploration has many important applications. However, classic information gain-based or frontier-based exploration only relies on the robot current state to determine the immediate exploration goal, which lacks the capability of predicting the value of future states and thus leads to inefficient exploration decisions. This paper presents a method to learn how ``good'' states are, measured by the state value function, to provide a guidance for robot exploration in real-world challenging environments. We formulate our work as an off-policy evaluation (OPE) problem for robot exploration (OPERE). It consists of offline Monte-Carlo training on real-world data and performs Temporal Difference (TD) online adaptation to optimize the trained value estimator. We also design an intrinsic reward function based on sensor information coverage to enable the robot to gain more information with sparse extrinsic rewards. Results show that our method enables the robot to predict the value of future states so as to better guide robot exploration. The proposed algorithm achieves better 
prediction and exploration performance compared with the state-of-the-arts. To the best of our knowledge, this work for the first time demonstrates value function prediction on real-world dataset for robot exploration in challenging subterranean and urban environments. 
More details and demo videos can be found at \url{https://jeffreyyh.github.io/opere/}.
\end{abstract}

\section{Introduction} \label{sec:intro}

\IEEEPARstart{I}{n} 
recent years, robot exploration has become more popular, ranging from search and rescue, space exploration, to the most recent DARPA Subterranean Challenge, where how to efficiently cover the unvisited areas and build the corresponding map is a key part to the success of the final task. 
Traditionally, information gain-based and frontier-based exploration methods are often used to select exploration goals by maximizing the information gain \cite{2002_Bourgault_InfoExploration} or unexplored frontiers \cite{1997_Yamauchi_frontier}. However, these methods only rely on the robot current state to determine the immediate  exploration goal, which are sensitive to the abnormal behavior of the robot and sometimes leads to the inefficient exploration decision. On the other hand, the longer horizon exploration history reflects the more comprehensive reason of the decision behind and can be better leveraged for exploring future unvisited area. This is especially critical for robot to explore large area or conduct repeated exploration for similar environments. 

More recently, 
reinforcement learning (RL)-based approaches \cite{NIPS2016_Bellemare} \cite{ICML2017_Pathak} \cite{2020_RIDE}  were used in robot exploration  \cite{2022ICRA_Bigazzi} \cite{2019ICLR_LearnExplo} \cite{2018_DRL_Exploation_Office} \cite{2020_explorationDRLgraph} \cite{2021_goalDriven_exploration_DRL}. \textcolor{black}{In these methods, the value functions and/or policy are learned from the experience sampled from the online interactions with the environment}. The learned value function can then be used to evaluate the current and future states of the robot, which guides the decision of the exploration. 
However, learning value function and/or policy via online interaction with the environment is not always practical for real-world robotic applications due to the costs of time and the potential safety concern.
Although we can generate synthetic data and train the policy in simulators, it may be challenging to transfer to real world due to the environment difference. 
OPE \cite{2015_DR} \cite{2021ICLR_benchmarkOPE} and offline RL \cite{2020_OfflineRLTutorial} \cite{2020NeurIPS_CQL} have become an emerging trend to tackle this problem by learning value function and/or policy via logged data to avoid online interaction with the environments.



\begin{figure}[]
  \centering
  \includegraphics[width=0.90\linewidth]{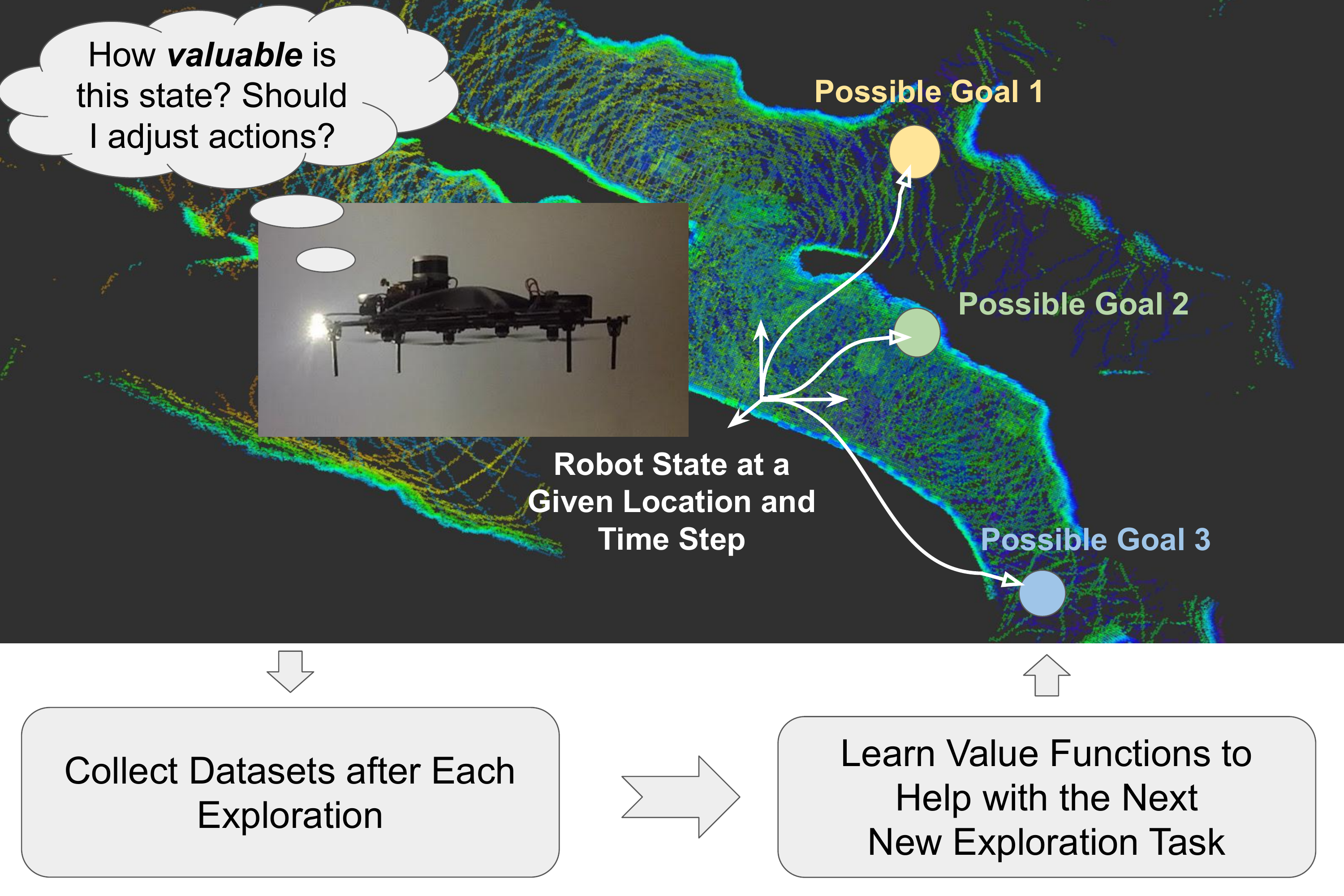}
    \caption{The robot explores the environments with a given policy $\pi$ and is able to decide how good\slash valuable the possible goals are. The states are evaluated by value functions following the policy $\pi$. We use real-world datasets collected under policy $\pi_B$ to learn the state value functions. The learned value function provides the guidance for the robot to make exploration decisions.
    \vspace{-2em}
    }
\end{figure}

There are few works investigating OPE on offline RL-based exploration problem \cite{2021ICLR_benchmarkOPE}. However, the existing benchmarks only use data from simple simulation environments such as MuJoCo. The application on real-world data especially in challenging environments has been less explored. 
One major challenge in OPE for robot exploration is the distributional shift between behaviour policy which collects the data, and the target policy which deploys online. Such distribution shift on exploration policy brings in challenge for the value function estimation in testing stage, because the learned value function approximator may no longer adapt to the distribution of the online testing data based on the target policy. In addition, the dataset from real world  adds extra difficulty to the OPE because the environment condition such as lighting, structure, etc is much more complex.

Another difficulty of the real-world exploration comes from the lack of extrinsic rewards, e.g. lack of environmental feedback. Sparse or even 
no extrinsic rewards 
make the exploration learning process inefficient \cite{2020_RIDE}. Although several intrinsic rewards were proposed to address this issue \cite{NIPS2016_Bellemare} \cite{2015_Stadie} \cite{ICML2017_Pathak}, most of them are only tested in simple non-robotic simulation environments. In real-world challenging subterranean or urban scenarios, the environments are highly unstructured, the sparsity of the extrinsic reward make the exploration even harder. Hence, a proper reward function design which can encourage the robots to acquire richer information about the environment is necessary to enable effective robot exploration.

In this paper, we propose to the learn value function for robot exploration problem. We formulate our work as the off-policy evaluation problem with offline Monte-Carlo (MC) training, where the datasets are collected in several challenging subterranean and urban environments. To tackle the distribution shift, we perform the Temporal Difference (TD) online adaptation scheme to optimize the offline trained value function approximator during testing. In particular, we design an ensemble of value function networks to further improve the estimation performance. Then, we design an intrinsic reward function based on sensor information coverage including both camera and LiDAR to enable the robot gain more information in the challenging environment with sparse extrinsic features. Different from other intrinsic reward designs which mainly rely on simulated camera information, our reward has a much richer representation of the real world which encourages the robot to explore the environment efficiently.

To summarize, the main contributions of this paper are:
\begin{itemize}
    
    \item We develop an OPE framework with offline MC training and TD online adaptation for robot exploration. To the best of our knowledge, this work for the first time demonstrates OPE with value function prediction for robot exploration in challenging environments. 
    \item We propose intrinsic rewards based on sensor information coverage which helps the robot acquire more information given sparse extrinsic rewards. 
    \item We design an ensemble of value networks to further improve the estimation performance.
    \item We release our datasets as an OPE benchmark for robot exploration in challenging environments.
    
\end{itemize}

\section{Related Work}

Classic exploration methods include information gain-based exploration \cite{ProbRobo} which selects actions based on a greedy strategy \cite{2002_Bourgault_InfoExploration} \cite{2016_Bai_InfoExploBayes} to maximize the information gain. Frontier-based methods are another common approach for robot exploration, where the frontiers are typically defined as some selected points in the boundary between explored and unexplored areas. In \cite{1997_Yamauchi_frontier}, the authors use a 2D occupancy grid map and detect frontier points by clustering the frontier edge segments. Other works such as \cite{2021_Batinovic_3DFrontier} use 3D occupancy grid maps and a more efficient frontier selection method. 
Although these traditional methods achieved success in robot exploration to some extent, these methods only rely on the current state to determine the immediate exploration goal, which 
may lead to inefficient exploration decisions.


Some intelligent methods were proposed for robot exploration, such as visitation count based\cite{NIPS2016_Bellemare}, curiosity-based \cite{ICML2017_Pathak}, memory based \cite{2019ICLR_EPISODIC}, etc. \cite{NIPS2016_Bellemare} \cite{ICML2017_Ostrovski} \cite{2017NIPS_Num_Exploration} use state visitation count as an intrinsic reward for exploration. Due to the high-dimensional continual state space, \cite{NIPS2016_Bellemare} \cite{ICML2017_Ostrovski} use a state pseudo-count model derived from Context-Tree Switching density model. \textcolor{black}{Curiosity-based intrinsic rewards \cite{2015_Stadie} \cite{ICML2017_Pathak} \cite{2022_CuriousExploration_WorldModels} were proposed to encourage agents to visited ``unexpected" states which have higher prediction error}. 
However, these algorithms usually require a large amount of samples to train the policy and value function in an online manner, which are sometime impractical to be deployed to real robots. 

More recently, off-policy evaluation (OPE) and offline reinforcement learning \cite{2020_OfflineRLTutorial} \cite{2020NeurIPS_CQL} algorithms have been used to train the policy and/or value function offline and then deployed online. OPE evaluates policies by estimating the value function of a target policy with data collected by a different behaviour policy to guide the decision making process \cite{2021ICLR_benchmarkOPE} \cite{2021NIPS_ES_OPE_RL} \cite{2016ICML_DROPE}. Although various OPE methods have been researched for different applications, few of them investigate the scenario on robotic exploration. In addition, the existing OPE methods \cite{2016ICML_DROPE} and benchmarks \cite{2021ICLR_benchmarkOPE} \cite{2021NIPS_ES_OPE_RL} rely heavily on data collected from simple simulation environments such as MuJoCo \cite{todorov2012mujoco}. 
The work in this paper is the first one to investigate off-policy policy evaluation for robot exploration in challenging environments. 

\section{Problem Formulation}\label{sec:pro}
The exploration procedure is formulated as a Partially Observable Markov Decision Process (POMDP) defined by the tuple $(\mathcal{O}, \mathcal{A}, \mathcal{R}, \mathcal{P}, \gamma)$.  $\mathcal{O} \in \mathbb{R}^m$ represents the observation space, which is the belief of the state space $\mathcal{S}$. $\mathcal{A} \in \mathbb{R}^n $ denotes the action space. $\mathcal{R}: \mathcal{S} \times \mathcal{A} \rightarrow \mathbb{R}$ is the reward space, $\mathcal{P}: \mathcal{S} \times \mathcal{A} \times \mathcal{S} \rightarrow \mathbb{R}_{+}$ denotes the stochastic state transition dynamic model, e.g., at time $t$, $p(s_{t+1}|s_t, a_t) \in \mathbb{R}_{+}$. We also define stochastic policy $\pi: \mathcal{O} \times \mathcal{A} \rightarrow \mathbb{R}_{+}$. The robot exploration trajectory $\xi$ is thus a tuple $\{o_0, a_0, r_0, \cdots, o_{T-1}, a_{T-1}, r_{T-1} \}$ following the MDP of the environment, with $o \in \mathcal{O}$, $a \in \mathcal{A}$, $r \in \mathcal{R}$. Here $T$ denotes the horizon of one exploration episode. The data used to learn the value function is a collection of trajectories, $\mathcal{D} = \{\xi_1, \xi_2, \cdots, \xi_M\}$ from a behavior policy. 
The state value function at time step $t$ given the exploration policy $\pi$ is thus formulated as the expected return starting from state $s$, where $\gamma \in [0, 1] $ denotes the discounting factor:
\begin{equation}
\begin{aligned}
    V_{\pi}(s) &= \mathbb{E}_{\pi, p} \left[ G_t | s_t = s \right ] \\
    &= \mathbb{E}_{\pi, p} \left[ \sum_{i=0} ^{T-1} \gamma ^ i R({t+i+1}) | s_t = s \right ]
\end{aligned}
\label{eq_Vs}
\end{equation}



\textcolor{black}{Off-policy evaluation aims at developing a value function estimator $\hat{V}_{\pi} ^{\mathcal{D}}$ (or simply $\hat{V}_{\pi}$, in our case, an approximate neural network model) given the historical data $\mathcal{D}$ collected by behaviour policy $\pi_{B}$ , so as to minimize the mean squared error (MSE) $\mathbb{E}[ (\hat{V}_{\pi}(\mathcal{D}) - V_{\pi} ]^2$.} In our off-policy evaluation problem setup, the goal is to learn the value function $\hat{V}_{\pi} (\phi(s))$ of the target policy $\pi(a|o)$, based on the training data collected from behaviour policy $\pi_B(a|o)$\footnote{\textcolor{black}{Our dataset was collected during the preparation for DARPA SubT challenge. The behavior policy during this period kept changing. Thus, the target policy that is for value function estimation is different from the policy that is for data collection (behavior policy).}}. We directly use the state observation $o \in \mathcal{O}$ as the state representation $\phi(s)$. 
The robot that is used to explore the environments is equipped with multiple sensors. Multiple information is considered when formulating the representations of the robot's state. Here, we use images captured from the on-board camera and the occupancy grid map as the state observations thus robot's state.
Thus, the state space $\mathcal{S}$ is continuous and we use a model-free function approximator to learn the state value function $V_{\pi}(s)$. Notice that the state transition $p(s_{t+1}|s_t, a_t)$ is unknown since the full structure and visual observations of the environment is unknown before the robot fully explores the environment. 


\section{Method}\label{sec:method}
This section describe our overall methodology. We first briefly introduce the environments. Then, we describe in detail the state representation and the intrinsic rewards design. Finally, the main value function approximation and the overall algorithm are presented.

\subsection{Maps and Environments}\label{subsec:map_env}
\begin{figure}
\vspace{-1em}
  \centering
  \includegraphics[width=0.75\linewidth]{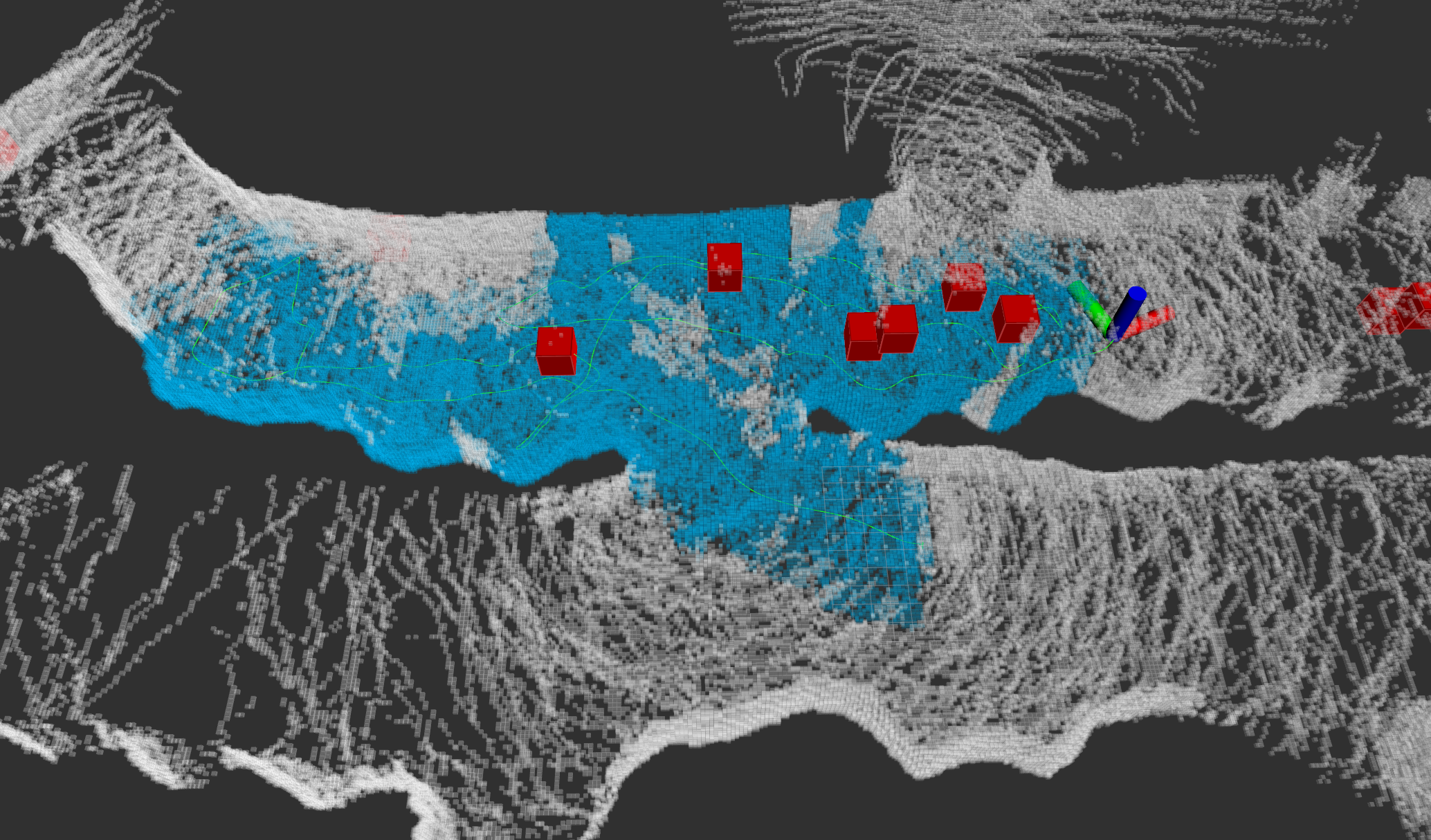}
    \caption{Illustration of the 3D map used for exploration. The light blue voxels denote the camera observed map, white voxels represent the frontier map; red boxes are the locations of the detected objects.}
    \vspace{-1em}
    \label{fig_mapRep}
\end{figure}

During exploration, we maintain two sets of maps: 1) a frontier map which contains regions which are not explored yet, and 2) a camera observed map which contains the regions within the frustum of the on-board camera for object detection. Both maps are represented as the 3D occupancy grid map. The reason we are interested in the camera observed map is that object discovery is one of the extrinsic rewards and thus more visual information means more chance of discovery interesting objects \cite{2020ECCV_VisualInterestingness}.
However the interesting objects are often sparse in the environments. The 3D map representation is shown in Fig. \ref{fig_mapRep}.

\subsection{State Representation}\label{subsec:state_rep}

As described in the Section \ref{sec:pro}, our state space $\mathcal{S}$ is continuous and affected by many factors, such as robot location, environment topology, visual information from the on-board camera, and the camera observed map and frontier map coverage. Here, we design the state representation to include two parts: (1) Visual information; (2) Local map around the robot. 

\begin{figure}
     \centering
     \begin{subfigure}[t]{0.235\textwidth}
         \centering
         \includegraphics[width=0.69\textwidth]{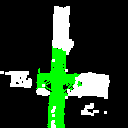}
         \caption{2D projected map image}
     \end{subfigure}
     \hfill
     \begin{subfigure}[t]{0.235\textwidth}
         \centering
         \includegraphics[width=0.69\textwidth]{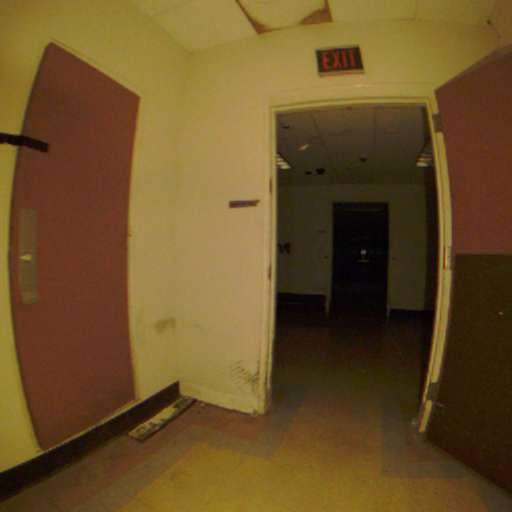}
         \caption{Image from on-board camera}
     \end{subfigure}
\caption{The state representation includes two parts: (a) the local 2D projected map from the 3D voxel map built by LiDAR; and (b) the RGB image from the on-board camera. In (a), the white pixels denote projected frontier map and the green pixels are camera observed map. }
\label{fig_stateRep}
\vspace{-1.5em}
\end{figure}

The visual information can be directly obtained from the RGB image captured by the on-board camera. We crop the image as a square to feed into the function approximator. As for the (2) local map, instead of simply cropping the local map and feeding the resulting 3D voxel map to a deep neural network-based function approximator, which is computational expensive, we use the 2D projection of the 3D occupancy map (including both camera observed map and frontier map), which also well preserves the shape of the local map as well as the map coverage. One example of the state representation is shown as Fig. \ref{fig_stateRep}.




\subsection{Intrinsic Rewards Design} \label{subsec:intrinsic_rewards}
To tackle the sparsity issue of the extrinsic reward in the challenging environment, we design an intrinsic reward function based on sensor information coverage to enable the robot gain much information. Our main goal is to let the robot gain as much visual information coverage and frontier coverage as possible and avoid visited regions while exploring the environment. Because visual information, including camera and LiDAR, directly affects object detection and semantic extraction. The unexplored regions (or frontier) on the map provides environments layout information. Both of them are usually vital to the exploration task.

To enable the robot continuously gather new information so that the total gained information increases overtime, we design the intrinsic reward based on the information increment between two consecutive time steps, which can be represented as the voxels on the map, either camera or frontier map. Overall, the reward is designed as: 
\begin{equation} \label{eq:reward}
    R(t) = a CG(t) + b LG(t) + c OG(t)
\end{equation}

\noindent where $CG(t) = C(t) - C(t-\Delta t)$ represents the camera visual coverage at time step $t$; $LG(t) = L(t) - L(t-\Delta t)$ is the LiDAR frontier map gain at time step $t$; $OG(t) = O(t) - O(t-\Delta t)$ expresses the extrinsic reward (new detected objects), although it is sparse. $C(t)$ and $L(t)$ denote the voxel number of camera observed map and LiDAR frontier map, respectively. $\Delta t$ is the time interval for computing the visual and LiDAR gains. $a$, $b$ and $c$ denotes the weight factors. 







Note that the intrinsic reward \eqref{eq:reward} is not used in the behaviour policy, which means the behaviour policy is not yet optimal for equation \eqref{eq_Vs}. The estimated value function can still be used for a new exploration task and serve as an heuristic for the planner used in the target exploration policy.


\subsection{Value Function Approximation} \label{subsec:funcAppro}
\begin{figure*}
  \centering
  \includegraphics[width=0.95\linewidth]{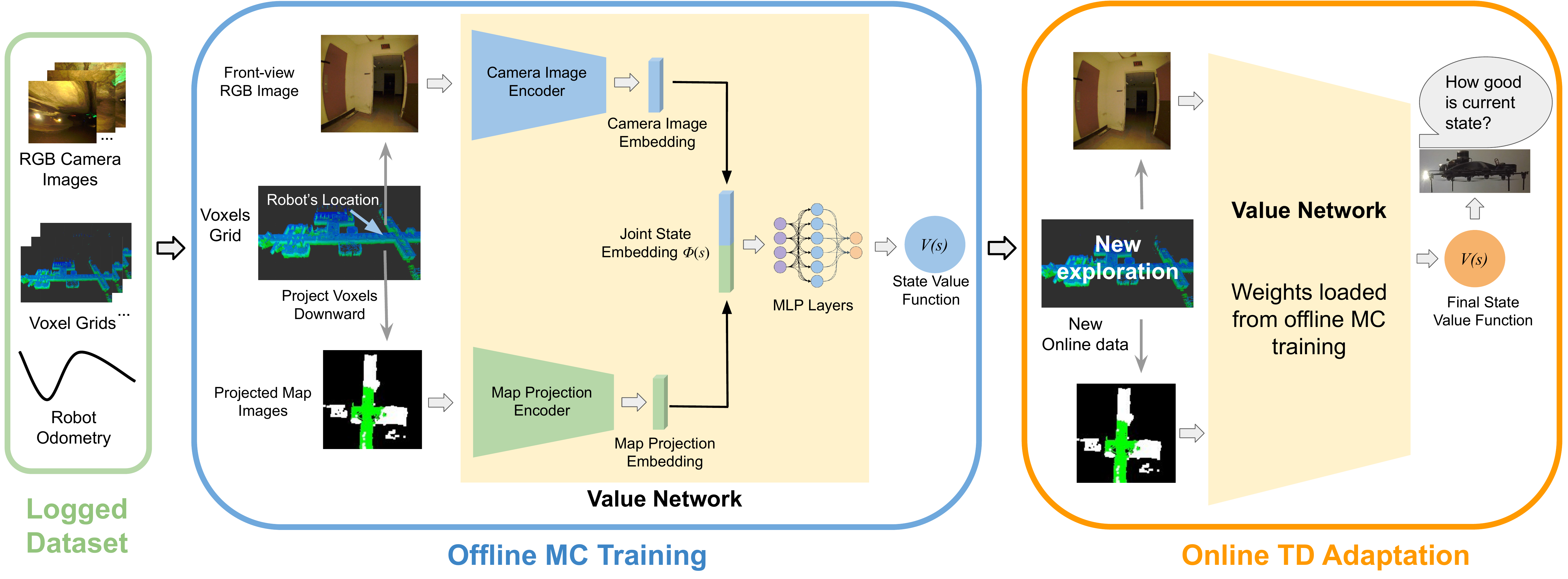}
\caption{Illustration of the value function approximation algorithm. First we collect datasets which consist of camera images and projected map images. Then we feed the data to the value function network described in Section \ref{value_net_structure} and perform offline MC pre-training, where the camera image and map projection image are sent to the encoders in parallel and then aggregated together to obtain the state value function. During the online deployment, we perform one additional TD adaptation step and get the final value function estimation for the robot to determine how good the current state is.}
\label{fig_algo_illu}
\vspace{-1.5em} 
\end{figure*}

The overall framework for the the proposed method is depicted in Fig. \ref{fig_algo_illu}. It consists of two major parts: offline MC pre-training and online adaptation using TD learning. The predicted value function tells how valuable the current state is for the robot, which can then guide the decision making for exploration.

\subsubsection{Offline pre-training and online adaptation}
The value function is approximated using a neural network with the the parameters denoted as $\boldsymbol{\theta}$. Thus, the prediction objective function can be formulated as,
\begin{equation}
    J(\boldsymbol{\theta}) = \sum_{s \in \mathcal{S}} \left[ V_{\pi}(s) - \hat{V}(s, \boldsymbol{\theta}) \right]^2
\end{equation}
where $V_{\pi}(s)$ and $\hat{V}(s, \boldsymbol{\theta})$ denote the true value function following policy $\pi$ and predicted value function, respectively.  

In the offline pre-training stage, we use Monte-Carlo (MC) method for value function approximation due to its low estimation bias \cite{RL_book}. The parameter updating rule is,
\begin{equation}
    \boldsymbol{\theta} = \boldsymbol{\theta} + \eta \left[G_t - \hat{V}(\phi(s), \boldsymbol{\theta}) \right] \nabla_{\boldsymbol{\theta}} \hat{V}(\phi(s), \boldsymbol{\theta})
    \label{eq_MC_train}
\end{equation}
where target value function under policy $\pi$ is approximated by the return $G_t$ of each training episode; $\eta$ denotes the learning rate.

In the online adaptation phase, we exploit the TD learning without waiting for the end of the current exploration episode to tackle the distribution shift. Bootstrapping of TD learning provides an extra opportunity to correct and improve the value function prediction based on the online data. Specifically, we apply a modified Bellman backup operator $\mathcal{B}$ repeatedly and get the recursive relationship of the state value function between two consecutive time steps $t$ and $t+1$.
\begin{equation}
    \mathcal{B} V(s_t) = R(t) + \gamma \mathbb{E}_{\pi, p} [V(s_{t+1})]
\end{equation}


Then, we use semi-gradient TD(0) to further update the parameters $\boldsymbol{\theta}$ and the overall updating rule is:
\begin{small}
\begin{equation}
\begin{aligned}
    \boldsymbol{\theta} &= \boldsymbol{\theta} + \eta \left[R(t) + \gamma \hat{V}(\phi(s_{t+1}), \boldsymbol{\theta}) - \hat{V}(\phi(s_t), \boldsymbol{\theta}) \right] \nabla_{\boldsymbol{\theta}} \hat{V}(\phi(s_t), \boldsymbol{\theta}) \\
    &= \boldsymbol{\theta} + \eta \left[ \mathcal{B} \hat{V}(\phi(s_t), \boldsymbol{\theta}) - \hat{V}(\phi(s_t), \boldsymbol{\theta}) \right] \nabla_{\boldsymbol{\theta}} \hat{V}(\phi(s_t), \boldsymbol{\theta})
\end{aligned}
\label{eq_TD_onlineLearn}
\end{equation}
\end{small}



\subsubsection{Value network structure} \label{value_net_structure}
The value function is approximated using a neural network. The network structure with the input state is illustrated in Fig. \ref{fig_algo_illu}.
Specifically, We use two encoders to encode the features of the camera image and 2D projected image. The encoded features are then concatenated and passed to a Multi-layer Perceptron (MLP) to get the final state value function prediction. In particular, we apply MobileNet-V3-Small \cite{MobileNetV3} model for both camera image encoder and map state image encoder considering the lower computational burden. 


\subsubsection{Value network ensemble}

Inspired by the approach in \cite{2018ICML_TD3}, we propose to train an ensemble of value function networks to reduce the prediction variance brought by offline MC training. The weights of these networks are denoted as $\boldsymbol{\theta}_1, \boldsymbol{\theta}_2, \cdots, \boldsymbol{\theta}_{N_V-1}$. $N_V$ denotes the number of value functions. Specifically, each individual value network will be trained offline and adapted online in the same way as described. Empirically, we found that the original estimator tends to overestimate the value function. Thus the minimum value of the ensemble is used as the estimated value function during online adaptation:
\begin{equation}
    \hat{V}(\phi(s), \boldsymbol{\theta}) = \operatorname*{min}_{\theta_i} \hat{V}(\phi(s), \boldsymbol{\theta}_i)
\end{equation}

\subsubsection{Overall algorithm}

The overall algorithms are then presented. Algorithm \ref{algo_offlineMC} describes offline pre-training with MC and Algorithm \ref{algo_onlineTD} describes online TD learning and testing. Note that we use $\boldsymbol{\theta}$ and $\boldsymbol{\theta}'$ to denote the network weights for training and testing, respectively, dataset $\mathcal{D}_{tr}$ and $\mathcal{D}_{te}$, etc. 

\begin{algorithm}[bht!]
\caption{Offline Pre-Training with MC}
\label{algo_offlineMC}
\KwInput{State representation $\phi(s)$ : camera image and cropped map}
\KwInput{Training trajectories dataset: $\mathcal{D}_{tr} = \{\xi_1, \xi_2, \cdots, \xi_M\}$}
\KwOutput{Learned value function: $\hat{V}(s, \boldsymbol{\theta}_1), \hat{V}(s, \boldsymbol{\theta}_2), \forall s \in \mathcal{S}$}



Initialize value networks weights $\boldsymbol{\theta}_1, \boldsymbol{\theta}_2$

\For{each training epoch}
{
    \For {$\xi_i \in \mathcal{D}$}
    {
        $T = $ length of $\xi_i$
        
        \For{$t = 0, 1, \cdots, T-1$}
        {
            Compute return as:
            $G_t = \sum_{i=0} ^T \gamma ^ i R({t+i+1})$
            
            Update parameters as :
            
            $\boldsymbol{\theta}_i = \boldsymbol{\theta}_i + \eta \left[G_t - \hat{V}(\phi(s_t), \boldsymbol{\theta}_i) \right] \nabla_{\boldsymbol{\theta}_i} \hat{V}(\phi(s_t), \boldsymbol{\theta}_i)$
            
            $\forall i \in \{1,2\}$
        }
    }
}
\end{algorithm}

\begin{algorithm}[bht!]
\caption{Online TD Adaptation}
\label{algo_onlineTD}
\KwInput{State representation $\phi(s)$: camera image and cropped map}
\KwInput{Pre-trained network weights: $\boldsymbol{\theta}_1, \boldsymbol{\theta}_2$}
\KwInput{Testing Trajectories data $\mathcal{D}_{te} = \{\xi\}$}
\KwOutput{Estimated value function $\hat{V} (s, \boldsymbol{\theta}')$}

\vspace{2mm} 

Initialize online network weights $\boldsymbol{\theta}'_1 = \boldsymbol{\theta}_1, \boldsymbol{\theta}'_2 = \boldsymbol{\theta}_2$

$T = $ length of $\xi$

\For{$t = 0, 1, \cdots, T-1$}
{
    Receive state observation $o_s(t)$
    
    Receive reward $R(t)$ following $\pi(a_t|s_t)$
    
    Transit to next state $s_{t+1}$ following  $\pi$ and $p(s_{t+1}|s_t, a_t)$
    
    Online update parameters as:
    
    $\boldsymbol{\theta}'_i = \boldsymbol{\theta}'_i + \eta \left[ \mathcal{B} \hat{V}(\phi(s_t), \boldsymbol{\theta}'_i) - \hat{V}(\phi(s_t), \boldsymbol{\theta}'_i) \right] \nabla_{\boldsymbol{\theta}'_i} \hat{V}(\phi(s_t), \boldsymbol{\theta}'_i)$
    
    $\forall i \in \{1,2\}$
    
    $\hat{V}(\phi(s_t), \boldsymbol{\theta}'_i) = \operatorname*{argmin}_{\theta'_i} \hat{V}(\phi(s), \boldsymbol{\theta}'_i)$
}
\end{algorithm}

\section{Experiments}

\subsection{Experimental Setup}
\subsubsection{Robot System}
\label{subsec:robotsystem}
\begin{figure}[ht!]
  \centering
  \includegraphics[width=0.95\linewidth]{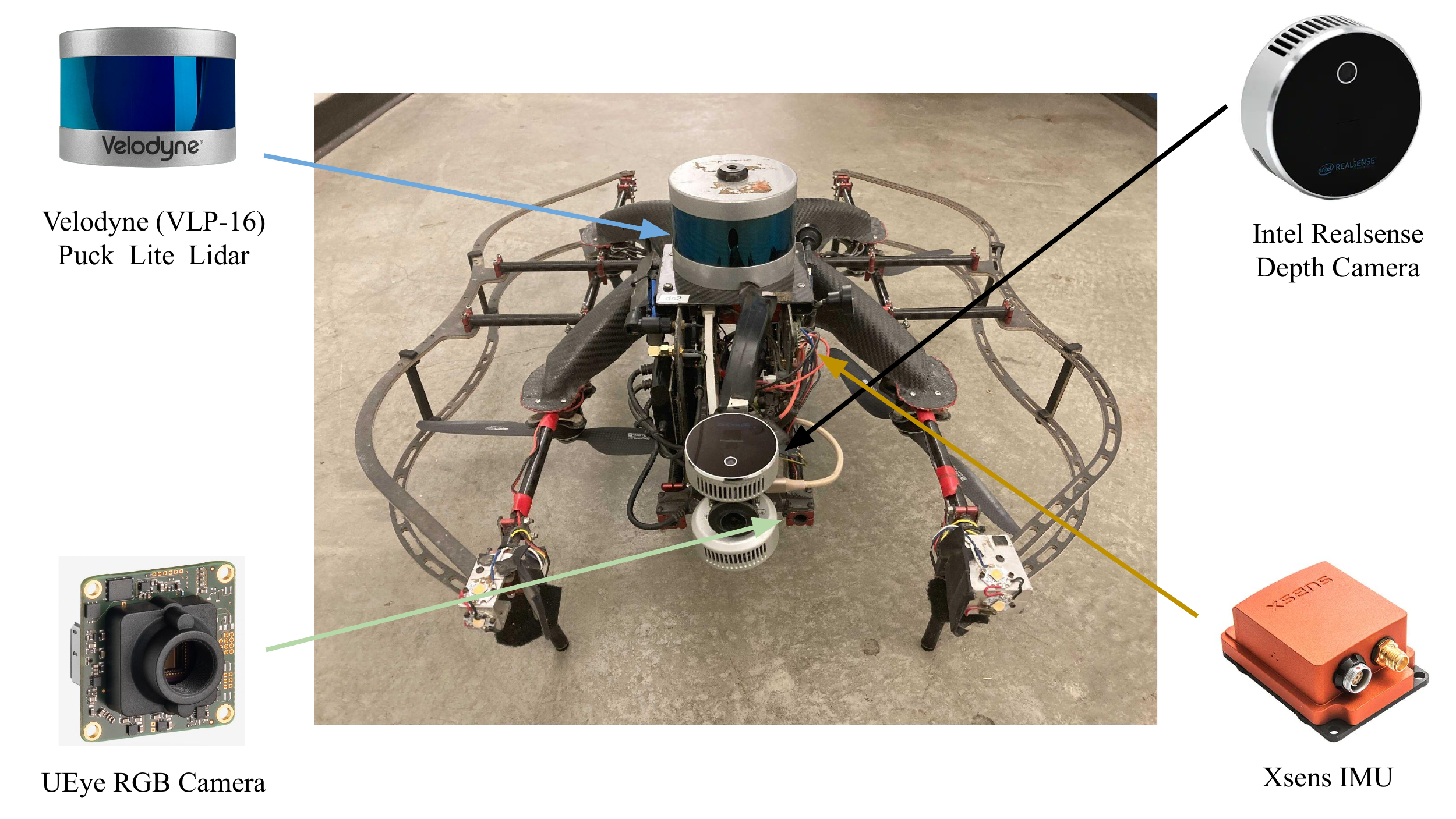}
\caption{The data collection platform. }
\vspace{-0.5em}
\label{fig_drone}
\end{figure}

The robot used for data collection during exploration is a custom-built quadcopter. It is equipped with a Velodyne (VLP-16) Puck Lite LiDAR, a Xsens MTi-200-VRU-2A8G4 IMU, a Intel Realsense L515, a UEye UI-3241LE-M/C RGB Camera and some wireless modules, as shown in Fig.  \ref{fig_drone}.

\subsubsection{Data Collection}
The exploration policy for data collection is based on a non-learning based frontier-exploration like algorithms, \textcolor{black}{which consists of a RRT based global planner, a local planner using A* and a trajectory generator that generates waypoints to the low-level controller. The global planner selects viewpoints sampled from the frontier based on a handcraft score which represents the weigheted sum of multiple factors}. \textcolor{black}{The global planner plans at 2Hz and the local planner plans at 10 Hz}. More details of the planning framework can be found in \cite{SubTUAV_2022}.
The drone explores several subterranean and urban environments, see Table~\ref{Tab:env}, which are similar to those in the DARPA Subterranean Urban Challenge \cite{SubT_2022}. In each exploration episode, we set the same starting point for the robot.
Each run varies from around 2 to 10 minutes depending on the size of the environments. For each environment, the number of exploration episode varies from 10 to 15 after selection. 




\begin{table}
\label{table_env}
\renewcommand{\arraystretch}{1.0}
\centering
\setlength{\tabcolsep}{0.5 mm}
{
\begin{tabular}{ c|c } 
\toprule
\textbf{Environment category} &\textbf{Descriptions}  \\
\midrule
 Auditorium corridor & structured urban indoor environment \\
Large open room &structured urban indoor environment \\
Limestone mine & long distance, wide open tunnels  \\
Natural cave & unstructured, narrow  \\
\bottomrule
\end{tabular}
\caption{Description of the environments for data collection.}
\label{Tab:env}
\vspace{-2em}
}
\end{table}

\subsubsection{Implementation details}
We use a pre-trained image-net as the initial network. The network is then trained for 50 epochs using the Adam optimizer with a learning rate of 0.0001. The discount factor is selected as $\gamma = 1.0$ for the accumulated reward. During online adaptation, we use the same learning rate. A two-network ensemble is used for value function estimation, $N_V = 2$. \textcolor{black}{As for the exploration policy, we use the learned model to predict a score (value function) for the candidate viewpoints sampled from the frontier map and select one as the exploration goal with the largest value to help the global planner make exploration decision.}

\begin{table*}[ht]
\renewcommand{\arraystretch}{1.0}
\centering
\begin{tabular}{c|cccc|cccc}
\toprule
Metrics &  \multicolumn{4}{c|}{Normalized RMSE ($\downarrow$ the lower the better)} & \multicolumn{4}{c}{R2 Score ($\uparrow$ the higher the better)}\\
\midrule
Methods & IS\cite{2021ICLR_benchmarkOPE} & FQE\cite{2021ICLR_benchmarkOPE}  & DICE\cite{2019NIPS_DualDICE}   & Ours
        & IS\cite{2021ICLR_benchmarkOPE} & FQE\cite{2021ICLR_benchmarkOPE}  & DICE\cite{2019NIPS_DualDICE}   & Ours \\
\midrule
Corridor Env.   &0.222$\pm$0.000 &0.192$\pm$0.000 &0.176$\pm$0.001 &\textbf{0.129$\pm$0.004}    
        &0.563$\pm$0.001 &0.672$\pm$0.001 &0.724$\pm$0.002
        &\textbf{0.853$\pm$0.010}   \\
Room Env.   & 0.398$\pm$0.001 &0.506$\pm$0.000 &0.512$\pm$0.000  &\textbf{0.213$\pm$0.002}    
      &-0.593$\pm$0.007 &-1.573$\pm$0.000 &-1.634$\pm$0.000  &\textbf{0.543$\pm$0.008}   \\
Mine Env. &0.272$\pm$0.000 &0.264$\pm$0.000 &0.282$\pm$0.001  &\textbf{0.207$\pm$0.002}    
      &0.064$\pm$0.003 &0.122$\pm$0.002 &-0.002$\pm$0.005  &\textbf{0.460$\pm$0.012}   \\
Cave Env. &0.535$\pm$0.000 & 0.532$\pm$0.000 &0.535$\pm$0.000  &\textbf{0.164$\pm$0.002}   
     &-2.234$\pm$0.000 &-2.198$\pm$0.000 &-2.235$\pm$0.000 &\textbf{0.695$\pm$0.006}  \\
\bottomrule
\end{tabular}
\caption{Value function prediction evaluated by NRMSE and R2 Score. Each entry contains the mean and std of three trails.}
\label{Tab:OPE}
\end{table*}

\subsection{Value function prediction evaluation}
We first compare with several methods in the state-of-the-art to demonstrate the advantages of the our method in value function prediction.

\noindent\textbf{Benchmark comparison} 

\noindent\textit{Importance Sampling (IS)}~\cite{2021ICLR_benchmarkOPE}, where in our implementation we use behavior cloning to get behaviour and target policies with training or testing data. We also use Monte-Carlo return to supervise the training similar to our proposed method. 

\noindent\textit{Fitted Q-Evaluation (FQE)}~\cite{2021ICLR_benchmarkOPE}, where the network is trained for value function estimation by bootstrapping from the action value function $Q(s , \pi(s))$. 

\noindent\textit{DICE}~\cite{2019NIPS_DualDICE, 2021ICLR_benchmarkOPE}, which does not require behavior policy for data generation or the direct usage of importance weights. We use BestDICE \cite{2021ICLR_benchmarkOPE} in our implementation.

\noindent\textbf{Evaluation metrics}
We report the following metrics:

\noindent\textit{Normalized RMSE (NRMSE)}: 
\begin{small}
\begin{equation}
    \text{NRMSE} = \frac{\text{RMSE}} {\hat{V}(s_t)_{max} - \hat{V}(s_t)_{min}}, \forall t \in [0, T-1]
\end{equation}
\end{small}

\noindent\textit{Coefficient of Determination (R2 score)}: measure the correlation of the predicted value function $\hat{V}(s)$ and ground truth value function $V_{\pi}(s)$. The highest value of R2 score is 1. 
\begin{footnotesize}
\begin{equation}
    \begin{aligned}
    R^2 = 1 - \frac{\sum_{t=0}^{T-1} \left[ \hat{V}(s_t) - V_{\pi}(s_t) \right]^2}{\sum_{t=0}^{T-1} \left[ \hat{V}(s_t) - \bar{\hat{V}}(s_t) \right]^2}, \bar{\hat{V}}(s_t) = \frac{1}{T} \sum_{t=0}^{T-1}\hat{V}(s_t)
\label{eq_r2}
\end{aligned}
\end{equation}
\end{footnotesize}

\begin{figure*}[]
\includegraphics[width=0.99\textwidth]{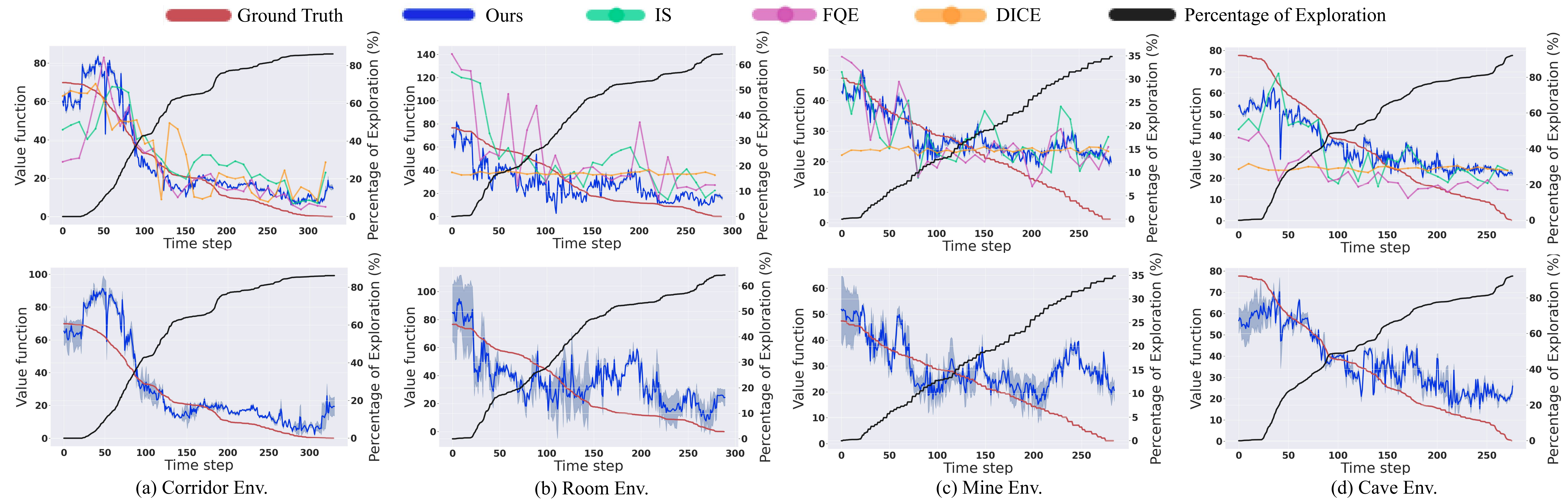}
\caption{Value function prediction performance and the comparison with baseline methods. Top row: the results of our method with model ensemble. Bottom row: the results with single value network. We plot the value prediction every 10 time step for other mehods. For our method, we plot value every step to show all the variance more clearly.}
\vspace{-1.0em}
\label{fig_pred_gt_V}
\end{figure*}

All methods under comparison are tested on four environments for value function prediction. 
Table~\ref{Tab:OPE} shows the evaluation results under two proposed metrics. Overall, thanks to the proposed offline Monte-Carlo training and TD online adaptation framework, our proposed method achieves the best performance with the lowest NRSME and the highest R2 score in all four environments, indicating that our method is better at predicating the value function. The importance-sampling method suffers from the inaccurate estimation of the importance sampling weights and thus cannot predict the value function very accurately, although sometimes can reduce the variance as expected. The prediction of FQE~\cite{2021ICLR_benchmarkOPE} and DICE~\cite{2019NIPS_DualDICE, 2021ICLR_benchmarkOPE} has the largest deviation from the ground truth due to the error accumulation in bootstrapping of the Bellman backup based method. In particular, the distribution mismatch between their behaviour and target policies makes this phenomenon more significant. 

\begin{figure*}
     \centering
     \begin{subfigure}[t]{0.49\textwidth}
         \centering
         \includegraphics[width=0.99\textwidth]{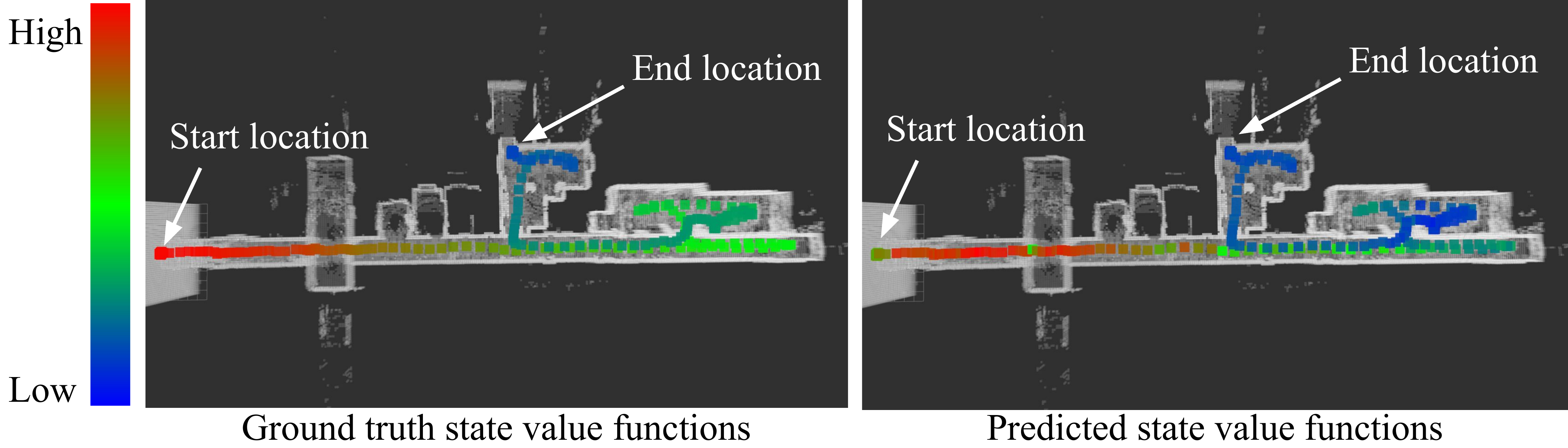}
         \caption{Corridor environment}
     \end{subfigure}
    \begin{subfigure}[t]{0.49\textwidth}
         \centering
         \includegraphics[width=0.99\textwidth]{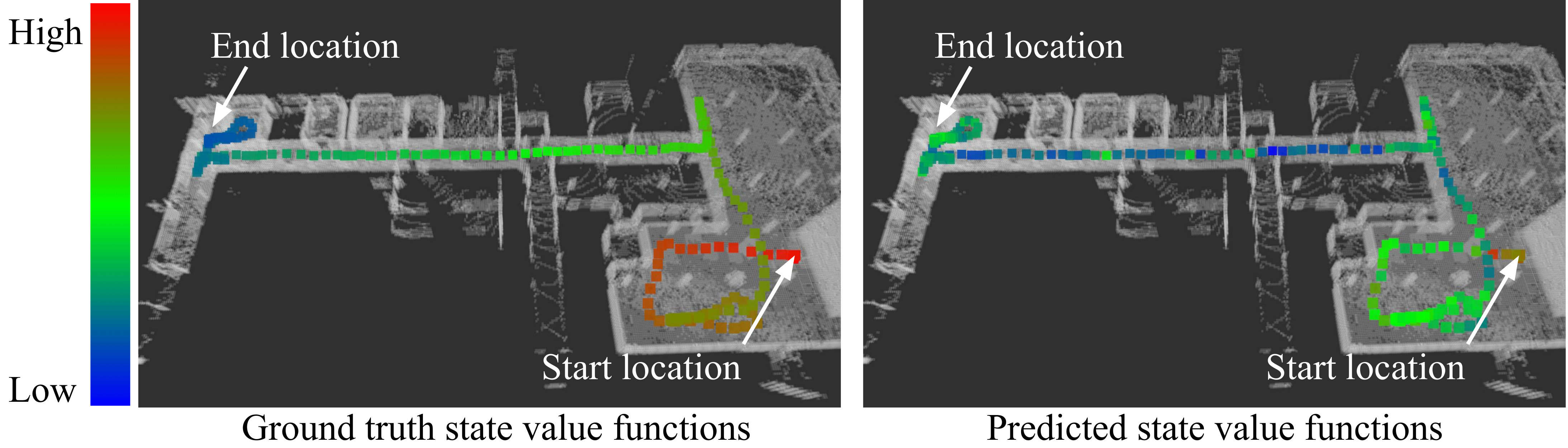}
         \caption{Room environment}
     \end{subfigure}
     \begin{subfigure}[t]{0.49\textwidth}
         \centering
         \includegraphics[width=0.99\textwidth]{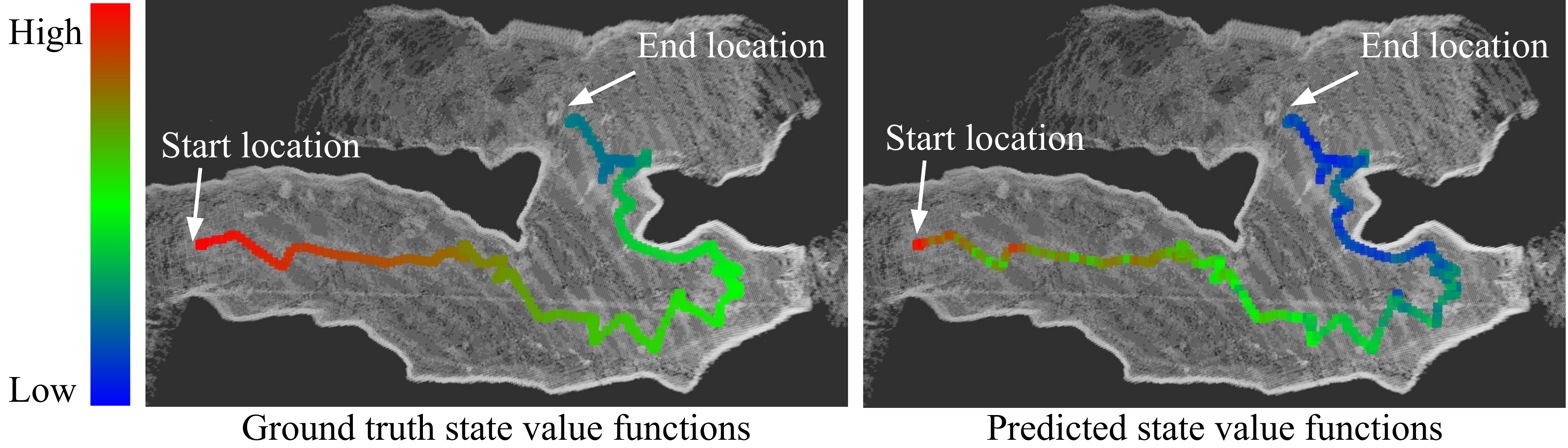}
         \caption{Mine environment}
     \end{subfigure}
     \begin{subfigure}[t]{0.49\textwidth}
         \centering
         \includegraphics[width=0.99\textwidth]{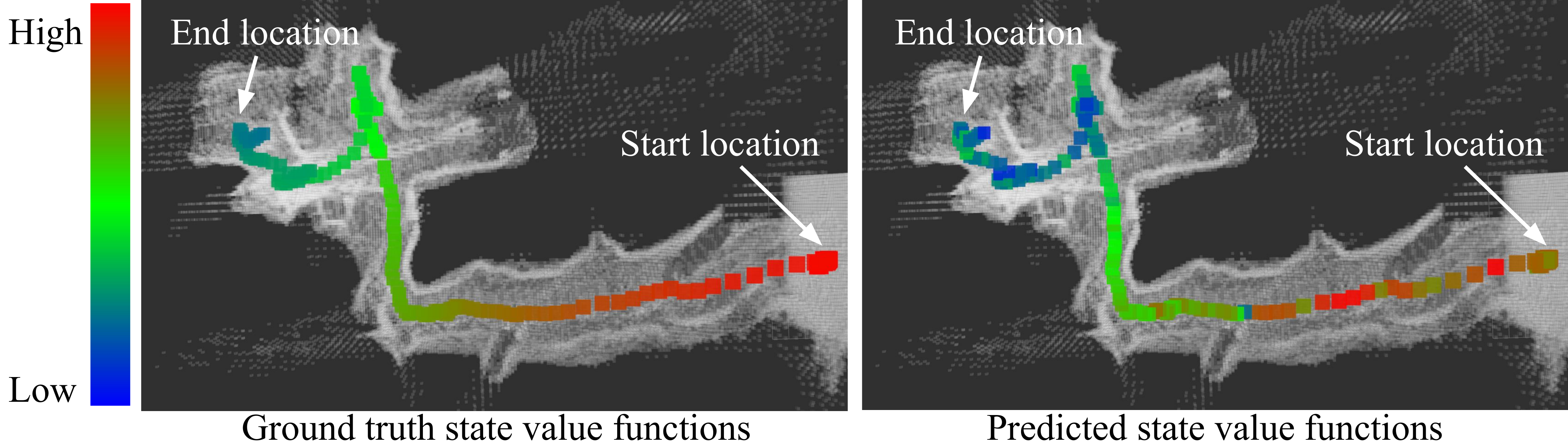}
         \caption{Cave environment}
     \end{subfigure}
\caption{Qualitative illustrations of the predicted value functions and ground truth value in different environments.}
\vspace{-1.5em}
\label{fig_quality_pred_gt_V}
\end{figure*}

Fig. \ref{fig_pred_gt_V} shows the time history of the value prediction performance. The percentage of exploration is defined as the voxel number of the camera observed map $N_{CM}$ over the voxel number of the global map $N_{GM}$, or $\varphi_{Explore} = N_{CM}/N_{GM}$. 
We also show the results of our method without using ensemble networks but just a single value network. 
Again, we can see that our method achieves better performance. Our prediction is the closest to the ground truth. It is also clear that using the ensemble networks can significantly reduce the prediction variance. 
Among the three baseline methods, importance sampling achieves better performance compared to the other two. The major reason is due to the use of Monte-Carlo return. 
We also present the qualitative results of the value function predictions during the robot exploration, shown in Fig.\ref{fig_quality_pred_gt_V}. 
We can see from both Fig.\ref{fig_pred_gt_V} and Fig.\ref{fig_quality_pred_gt_V} that when the exploration starts, the robot predicts very high value function indicating that there is still lots of new information to gain at the current state for the whole environment exploration. As the exploration process going, the predicted value becomes lower and gets to the lowest in the end. This is because as the robot gradually completes the exploration task, the unexplored region becomes less and less leading to less remaining new information to obtain and thus low predicted value function. 

\subsection{Regret evaluation}
\begin{figure}[]
  \centering
  \includegraphics[width=0.80\linewidth]{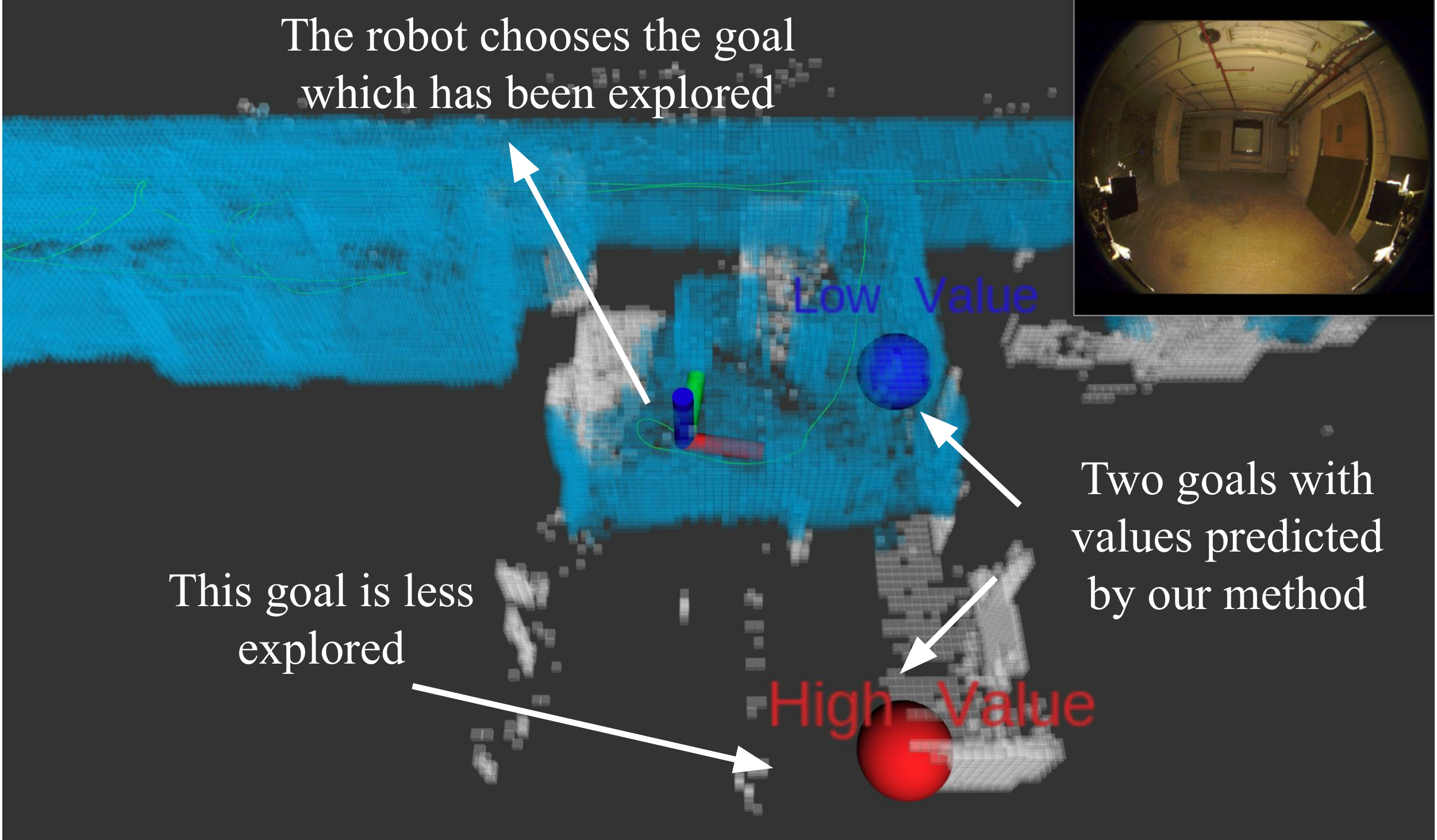}
\caption{The red goal is less explored and denotes higher predicted value while the blue target has been fully explored indicting lower value if robot pursues this goal. Human expert picks the red target as the \textit{correct} goal. The robot selects the blue goal and made the \textit{incorrect} decision in this scenario.}
\vspace{-2em}
\label{fig_value_decision}
\end{figure}

To show the benefit of the proposed approach during exploration, we compare the correctness of the decision for robot exploration of our method with the three OPE baselines using learned value function as well as frontier-based exploration.
When using the learned value function for exploration, the robot always selects an exploration goal with a higher estimated value. With the frontier based policy, the logged actual decision was based on the frontier exploration policy, which may not choose goals with higher value.
The correct decisions (or ground truth) is generated by the human expert based on the state at that time step\footnote{\textcolor{black}{Due to the global perspective, humans make better decisions (can be approximated as optimal) to select more informative exploration goals compared to the robot which only had instantaneous online information.}}.
Fig. \ref{fig_value_decision} shows the correct or incorrect decision at some key location. The blue voxels are regions with visual coverage while the white voxels are LiDAR map frontiers. 

We use regret \cite{OLOCO} to measure the correctness of the decisions. Specifically, we count the mistakes made by the robot using different methods.

\vspace{-1.5em}
\begin{equation}
    \text{Regret} = \sum_{t=0}^{T-1}  \left[ R_C(\pi^*(a_t|o_t)) - R_C(\pi(a_t|o_t)) \right]
    \vspace{-0.5em}
\end{equation}

where $R_C$ denotes correct exploration decision.  $\pi^*(a_t|o_t)$ is the policy from the human expert and $\pi(a_t|o_t)$ is the policy under compared (high value function or non-learning based policy). Smaller regret indicates less mistakes. 

Table.~\ref{Tab:Reg} presents the results of the regret evaluation\footnote{We selected 12 decision points for the corridor env., 7 for the room env., 8 for the mine and cave env. The corridor env. has more bifurcated hallways and thus more key points to make decisions. We run 10 i.i.d. tests.}. It is clear that our method achieves the best performance with the least mistakes. The robot made better decisions using the accurate learned value function as the guidance because it is able to leverage the information gained in the exploration history. The results shows that a learned value function enables the robot to predict how valuable future states are so that it is possible to better guide the decision making for robot exploration.

\vspace{-0.5em}

\begin{table}[ht!]
\color{black}
\renewcommand{\arraystretch}{1.0}
\centering
\resizebox{\columnwidth}{!} 
{
\begin{tabular}{c|cccc}
\toprule
Methods & Corridor Env. &Room Env. & Mine Env. &Cave Env. \\
\midrule
Frontier \cite{SubTUAV_2022}  &0.333 &0.714 &0.625 &0.250 \\
IS \cite{2021ICLR_benchmarkOPE}   &0.633 &0.443 &0.363 &0.625 \\
FQE \cite{2021ICLR_benchmarkOPE} &0.367 &0.286 &0.263 &0.263 \\
DICE \cite{2019NIPS_DualDICE}  &0.617 &0.343 &0.125 &0.375 \\
Ours &\textbf{0.108$\pm$0.038} &\textbf{0.114$\pm$0.057} &\textbf{0.100$\pm$0.050} &\textbf{0.013$\pm$0.038} \\
\bottomrule
\end{tabular}
}
\caption{\textcolor{black} {Normalized regrets in different environments. Lower regret means better decisions made by the agent. Since our method involves network ensemble, we also show the mean and variance.}} 
\label{Tab:Reg}
\vspace{-2.0em}
\end{table}

\subsection{Real Robot Experiment for Exploration}
\begin{figure}[ht!]
\vspace{-0.5em}
  \centering
  \includegraphics[width=0.95\linewidth]{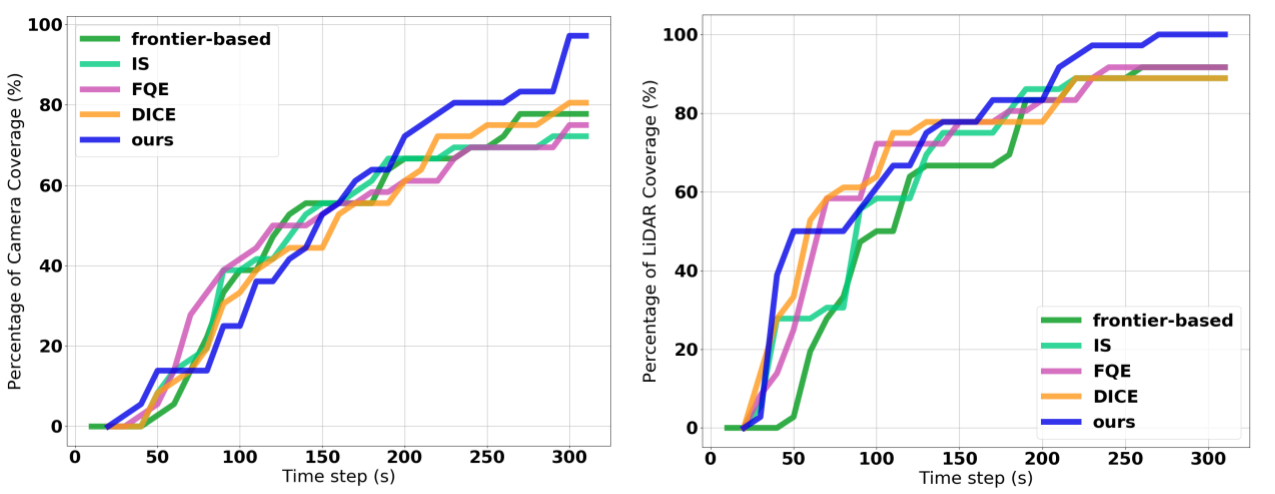}
\caption{Performance measured by camera and LiDAR coverage.}
\vspace{-1.5em}
\label{fig_explor_performance}
\end{figure}

We then tested our method on the same robot in the corridor environment for the real exploration task. We use the same planner as in \cite{SubTUAV_2022}. 
We directly deploy our trained model to predict the viewpoints scores.
\textcolor{black}{
We evaluate the exploration performance with camera and LiDAR coverage. 
Again, we compare the exploration performance with other baselines, see Figure~\ref{fig_explor_performance} and Table \ref {Tab:exp_cover}. 
}
\textcolor{black}{Since multiple value function predictions are conducted at each run, the results can reflect the statistical significance of the value estimation for exploration.}
\textcolor{black}{Results show that by using our method, the robot always selects the viewpoints with higher scores and thus explore higher value regions. The traditional frontier-based method \cite{SubTUAV_2022} cannot always make the correct decision of the exploration goals due to the hand craft viewpoint scores. Due to the inaccurate estimation of the important sampling weights and the error accumulation in bootstrapping, the other three baselines cannot predict the value function very accurately. In fact, all the baseline methods ended with more repeated paths and therefore inefficient exploration performance in the real flight tests.} Our method can make the robot successfully explore the whole environment efficiently. 



\begin{table}[ht!]
\color{black}
\renewcommand{\arraystretch}{1.0}
\centering
\resizebox{\columnwidth}{!} 
{
\begin{tabular}{c|ccccc}
\toprule
Methods & Frontier-based \cite{SubTUAV_2022}  &IS \cite{2021ICLR_benchmarkOPE} & FQE \cite{2021ICLR_benchmarkOPE}  & DICE \cite{2019NIPS_DualDICE}  &Ours \\
\midrule
Camera coverage  &77.78\% &72.22\% &75.00\% &80.56\% &\textbf{97.22}\% \\
LiDAR coverage  &91.67\% &88.89\% &91.67\% &88.89\% &\textbf{100}\% \\
\bottomrule
\end{tabular}
}
\caption{\textcolor{black}{Camera and LiDAR coverage at the end of episodes}} 
\label{Tab:exp_cover}
\vspace{-2.0em}
\end{table}

\subsection{Ablation Study}
\textcolor{black}{An ablation study investigates the the effectiveness of our online adaption using TD learning is presented in Table \ref {Tab:Abla}.}

\textcolor{black}{We compare the results with/without using TD learning for online adaptation. We noticed that without TD learning, the value prediction performance deteriorates in most cases. The only exception is the cave environment. Cave environment has smaller amount of data thus the model overfit to the training data if using the same setting for other environments. This makes TD learning less effective comparing with in other environments. TD learning effectively made the correctness to adapt to the online environment. The overall method outperforms the baselines by a large margin.
}

\vspace{-1.0em}

\begin{table}[ht!]
\color{black}
\renewcommand{\arraystretch}{1.0}
\centering
\resizebox{\columnwidth}{!} 
{
\begin{tabular}{c|cc|cc}
\toprule
Environment &  \multicolumn{2}{c|}{Normalized RMSE ($\downarrow$)} & \multicolumn{2}{c}{R2 Score ($\uparrow$)}\\
{}   & w/o TD   & w/ TD  
        & w/o TD    & w/ TD \\
\midrule
Corridor   &0.143$\pm$0.000  &\textbf{0.129$\pm$0.004}    
            &0.818$\pm$0.001  &\textbf{0.853$\pm$0.010}   \\
Room   &0.258$\pm$0.000   &\textbf{0.213$\pm$0.002}    
      &0.333$\pm$0.001   &\textbf{0.543$\pm$0.008}   \\
Mine   &0.261$\pm$0.001  &\textbf{0.207$\pm$0.002}    
      &0.271$\pm$0.005   &\textbf{0.460$\pm$0.012}   \\
Cave  &\textbf{0.155$\pm$0.000}  &0.164$\pm$0.002  
      &\textbf{0.727$\pm$0.001}  &0.695$\pm$0.006   \\
\bottomrule
\end{tabular}
}
\caption{\textcolor{black} {Ablation Study on Online Adaptation.}}
\label{Tab:Abla}
\vspace{-1.5em}
\end{table}

\section{Conclusions and Future Work}
In this paper, we present a method for predicting state value function from logged real-world data to guide the robot exploration. Our method consists of offline MC training and online TD adaptation. A coverage-based intrinsic reward is designed to encourage exploration. The proposed method was validated on various challenging environments. Results show that learned value function enables the robot to predict the value of future states to better guide the decision making for exploration. Our approach outperforms the state-of-the-art under all environments.  
In the future, we plan to further incorporate this value function prediction with the planner to improve the exploration policy.

{\footnotesize
\bibliographystyle{unsrt} 
\bibliography{main}
}

\end{document}